\documentclass[final]{cvpr}

\usepackage{times}
\usepackage{epsfig}
\usepackage{graphicx}
\usepackage{amsmath}
\usepackage{amssymb}
\usepackage{multirow}
\usepackage{adjustbox}
\usepackage{booktabs}
\usepackage{bbding}
\usepackage{subcaption}
\usepackage[font=small]{caption}
\usepackage{tabularx}
\usepackage{array}

\newcommand{\R}{\mathbb{R}}
\usepackage[pagebackref=true,breaklinks=true,colorlinks,bookmarks=false]{hyperref}

\def\cvprPaperID{3} 
\def\confYear{CVPR 2021}

\begin{document}

\title{Occlusion Guided Scene Flow Estimation on 3D Point Clouds}

\author{Bojun Ouyang\\
Tel Aviv University\\
{\tt\small bojungouyang@mail.tau.ac.il}
\and
Dan Raviv\\
Tel Aviv University\\
{\tt\small darav@tauex.tau.ac.il}
}

\maketitle

\begin{abstract}
3D scene flow estimation is a vital tool in perceiving our environment given depth or range sensors. Unlike optical flow, the data is usually sparse and in most cases partially occluded in between two temporal samplings. Here we propose a new scene flow architecture called OGSF-Net which tightly couples the learning for both flow and occlusions between frames. Their coupled symbiosis results in a more accurate prediction of flow in space. Unlike a traditional multi-action network, our unified approach is fused throughout the network, boosting performances for both occlusion detection and flow estimation. Our architecture is the first to gauge the occlusion in 3D scene flow estimation on point clouds. In key datasets such as Flyingthings3D and KITTI, we achieve the state-of-the-art results.\footnote{Our code will be publicly available upon publication.}
\footnote{https://github.com/BillOuyang/OGSFNet.git}

\thispagestyle{empty} 


\end{abstract}

\section{Introduction}

Scene flow estimation is a core challenge in computer vision which aims to find the 3D motion between points from consecutive temporal frames. While flows in between images, also known as optical flow, still have an important part in modern vision systems, the rise of depth sensors shifts the focus towards geometric flows. The two tasks are similar in spirit but with one fundamental gap - the data source for optical flow are regular dense samples given on top of a grid, while most depth sensors, especially outdoor, provide a sparse set of points in space. Algorithmic-wise, in the deep networks era, that gap shifts us from image-based convolutions towards graph neural network architectures.

Early attempts to solve 3D model alignment minimized the point-to-point or point-to-plane energy and were referred to as Iterative-Closest-Point (ICP) algorithms~\cite{10.1117/12.57955,chen_medioni}, where during iterative steps one searches for the closest set of matched points and minimizes the energy on that subset. Rigid alignment ~\cite{10.1117/12.57955} was first introduced, then rapidly non-rigid deformations were solved by adding adequate regularization~\cite{4270190}.
Many different approaches for alignment appeared over the years. Just stating a few - ~\cite{huguet_devernay_2007,cech_sanchez-riera_horaud_2011,ilg_saikia_keuper_brox_2018,herbst_ren_fox_2013,gottfried_fehr_garbe_2011} focused on RGB-D between stereo images, ~\cite{raviv_bronstein_bronstein_kimmel_sochen_2011} introduced a more robust cost function, ~\cite{raviv_dubrovina_kimmel_2012} considered the alignment as a quadratic assignment task and ~\cite{doi:10.1137/120888107} added intrinsic long geodesics to enrich the process with global features. 

Moving from axiomatic methods towards learning-based approach became feasible lately, where graph convolutions evolved as well as rich and deep enough networks were capable of sensing more of the scene. 
FlowNet3D~\cite{liu:2019:flownet3d} was probably the first robust learnable deep network for aligning 3D point clouds. It utilizes a PointNet++~\cite{qi2017pointnetplusplus} structure and computes the correlation between the point clouds by using the flow embedding layer. Following that line of thought PointPWC-Net~\cite{wu2020pointpwc}, based on an optical flow mechanism ~\cite{Sun2018:Model:Training:Flow}, uses Feature Pyramid Network on top of local correlations with a new cost volume and cost function, showing superior results all across the benchmarks. Lately, we have seen attempts to handle larger sets of deformations by correlating all points with all points~\cite{puy20flot} but those models require a massive increase in memory resources and suffer from outliers that need to be cleaned.

\begin{figure}[t]
\begin{center}
\includegraphics[width=0.5\textwidth]{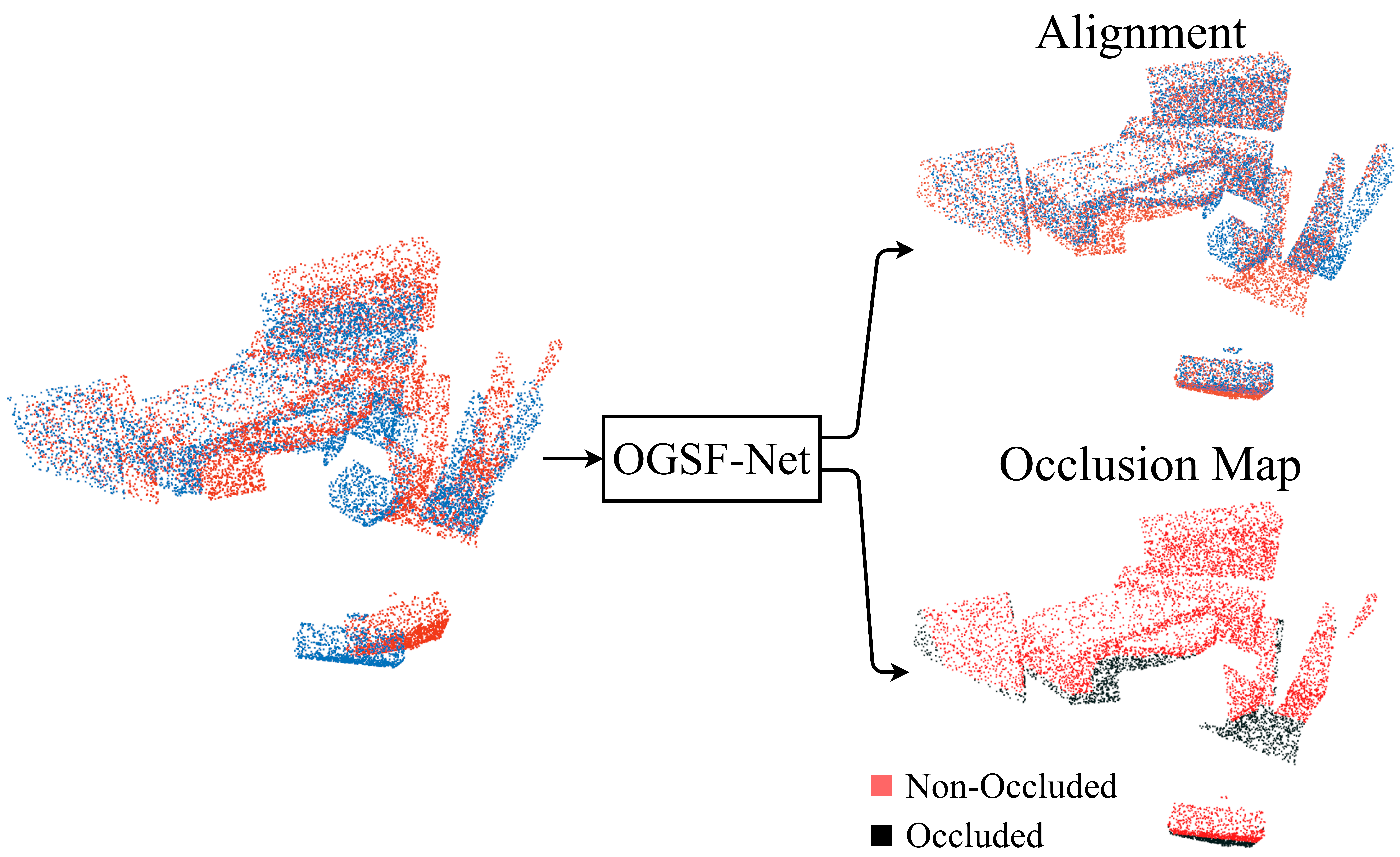}
\end{center}
    \vspace{-15pt}
   \caption{\textbf{Multi-task model}. OGSF-Net directly consumes the point clouds from  two different frames as its input. It predicts the scene flow and occlusion map of the source relative to the target. 
   }
\label{fig:teaser}
\end{figure}

When calculating flow in between objects, we encounter in many cases the challenge of occlusions, where some regions in one frame do not exist in the other. Due to the displacement between the sensor and the object, the sensor does not see the entire object in all time steps. Incorrect treatment of the occluded area would reduce the performance of the flow estimation. That is true for optical flow tasks in images and of course for scene flow.
Classical methods usually regularize incoherent motion to propagate flow from non-occluded pixels to the occluded region ~\cite{saxena2019pwoc,luo2020occinpflow}. That is also true in the deep learning era where occlusions were learned in addition to flow estimation. Those attempts worked well on regular grids but traditionally failed on a sparse set of points due to numerical challenges. In this work, we focus on that exact task and show for the first time that if we couple the task for flow and occlusion tight enough in a guided approach we can gain in both worlds; getting a more accurate flow and understanding what is occluded.

\vspace{0.5cm}
\noindent
The main contributions of our work are:
\begin{itemize}
\item We propose a deep learning model called OGSF-Net which can jointly estimate the scene flow and occlusion map from point clouds.
\item  We utilize an occlusion handling mechanism inside our Cost Volume layer.
\item We present a new residual multi-scale architecture in place of traditional multi-scale flow schemes. 
\item  We show state-of-the-art performance on Flyingthings3D and KITTI Scene Flow 2015.

\end{itemize}


\section{Related Work}

\noindent
\textbf{Deep Learning on Point Clouds}. 
Deep learning has been proved to be one of the most successful learning tools in image processing and swept the community towards new achievements over axiomatic modelling. Graph neural networks, focusing on a more generalized structure, where vertices and edges represent our data, followed the revolution, presenting exciting new tools to handle irregular data. In computer vision, point cloud is one very common way to represent geometry acquired by range sensors or generated in virtual worlds. We have seen papers coping with new challenges by sampling the data and projecting the points into volumetric lattices ~\cite{wu_song_khosla_yu_zhang_tang_xiao_2015, allen_2010, qi_su_niebner_dai_yan_guibas_2016,liu2019pvcnn} and later on focusing on point convolutions or a combination of edge and points pulling layers, known as message passing ~\cite{Wang2019_GACNet,charles_su_kaichun_guibas_2017, qi2017pointnetplusplus,ravanbakhsh2016deep,su_jampani_sun_maji_kalogerakis_yang_kautz_2018,tatarchenko_park_koltun_zhou_2018,su18splatnet,hua_tran_yeung_2018,accv2018/Groh,verma_boyer_verbeek_2018,li2018pointcnn,liu2019pvcnn}.
Interesting follow-up papers appeared rapidly, trying to solve the main challenge arising from the permutation challenge in graphs. We do see recently different sampling strategies or different pulling methods. MLP layers and MAX-pooling are two relevant and popular building blocks for that ~\cite{charles_su_kaichun_guibas_2017,ravanbakhsh2016deep}.
Another interesting and popular approach was using the points as raw data input ~\cite{charles_su_kaichun_guibas_2017}, followed by a hierarchical architecture which can capture the local structure of the point clouds ~\cite{qi2017pointnetplusplus}.
Treating point cloud as graph and performing convolution over local neighbourhood make a lot of sense, and several successful approaches were introduced lately focusing on the convolution engine ~\cite{brab2016dynamic, simonovsky_komodakis_2017, wang_suo_ma_pokrovsky_urtasun_2018,Hermosilla_2019,dgcnn, wu_qi_fuxin_2019}. In our work, we use the PointConv suggested by~\cite{wu_qi_fuxin_2019,wu2020pointpwc} to perform the convolutions on the point clouds.

\noindent
\textbf{Scene Flow Estimation on Point Clouds}.
The increasing popularity of range data gave birth to the need for fast and accurate mapping of point clouds. 
\cite{7759282,behl_paschalidou_donne_geiger_2019, rishav2020deeplidarflow,7989666} suggested estimating the scene flow directly from real LiDAR scans.~\cite{7759282,behl_paschalidou_donne_geiger_2019}  consider the scene flow as a rigid motion, while \cite{wang_suo_ma_pokrovsky_urtasun_2018,liu:2019:flownet3d,gu_wang_wu_lee_wang_2019,wu2020pointpwc,liu2019meteornet,puy20flot} remove those restrictions.
Based on the~\cite{qi2017pointnetplusplus} architecture, FlowNet3D~\cite{liu:2019:flownet3d} introduced a novel flow embedding layer that aggregates the features from different frames. However, they only applied the flow embedding at a certain scale which limits the allowed feasible gap between the frames.~\cite{wu2020pointpwc} introduced a neural network based on \cite{Sun2018:Model:Training:Flow} which can predict the scene flow in a coarse-to-fine manner, showing superior results both for large and small flows. However, they do not have any treatment for the occlusions, and their accuracy  decrease significantly when there are occluded regions in the point clouds. 
Recently,~\cite{DeepLiDARFlow2020} suggests estimating the scene flow using both RGB and LiDAR data to overcome ambiguity by providing an additional layer of information. ~\cite{puy20flot} proposed an interesting approach focusing on all-to-all correlation using graph matching.

\noindent
\textbf{Occlusion estimation in Scene Flow}. Scene flow estimation and occlusion are treated as a chicken-and-egg problem as they are highly related to each other and one influence the other. Many papers~\cite{Hur:2019:IRR,ilg_saikia_keuper_brox_2018,Janai2018ECCV,saxena2019pwoc} suggest to predict the occlusion mask jointly with the flow and to refine the flow estimation by using the predicted occlusion mask. ~\cite{hur_roth_2017,Wang_2018_CVPR,Meister:2018:UUL} suggest to predict both the forward and backward flow and to find the occluded region based on the warped images. In ~\cite{Janai2018ECCV}, they proposed an unsupervised training framework which can predict the optical flow and occlusion from multiple frames. Based on ~\cite{Sun2018:Model:Training:Flow}, PWOC-3D~\cite{saxena2019pwoc} suggests a self-supervised strategy for the occlusion estimation by masking the warped feature inside the Cost Volume layer using the occlusion map.

In this work, we suggest entangling the two aspects together all across the network, not just only in the cost function. We claim that the occlusion should guide the flow and vice-versa as part of the architecture itself to gain the most out of the two. To the best of our knowledge, we are the first to estimate the occlusion in 3D scene flow estimation on point clouds and the first to present a guided linked unit in the pipeline to solve the flow-occlusion coupled task. We present state-of-the-art alignment results over all methods described above on known datasets.

\begin{figure*}
\begin{center}
\includegraphics[width=1.0\textwidth]{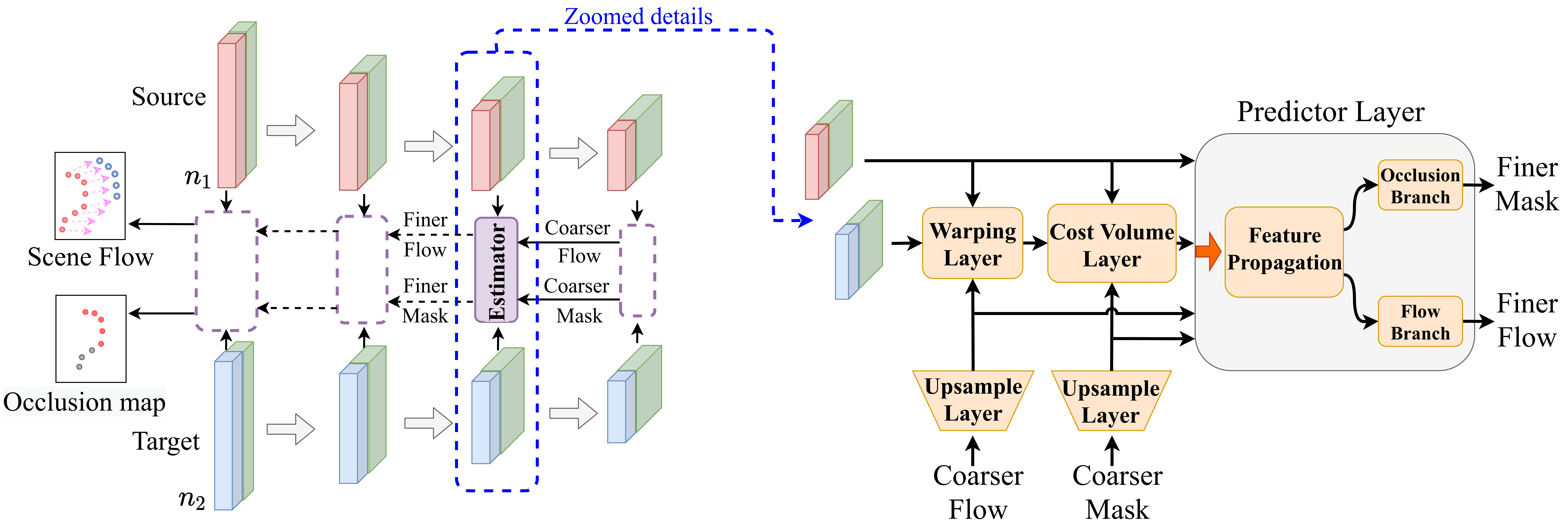}
\end{center}
    \vspace{-20pt}
   \caption{\textbf{Architecture}. On the left, we show the entire pipeline of OGSF-Net. It takes the input point cloud on the left and uses the PointConv+FPS to downsample the point cloud at each level. On the right, we show the finer details at each level. We first warp the target towards the source in order to construct our Cost Volume. Using the PointConv and MLP in the Feature propagation layer, we create the shared input features for the Flow/Occlusion branch.} 
\label{fig:full_arch}
\end{figure*}

\section{Problem Definition}
Given two samplings of a 3D scene we wish to estimate the movement in space between the source and the target and to identify the points in the source that do not appear in the target.
We represent the two sampled scenes, source $S$ and target $T$ as point clouds. Specifically,
$S=\{p_i|p_i\in \R^3\}_{i=1}^{n_1}$ with $n_1$ points and $T=\{q_j|q_j\in\R^3\}_{j=1}^{n_2}$ with $n_2$ points. Each point can also have a feature vector such as color or local normal to the surface. To ease notations, we denote $c_i\in\R^d$ the feature of point $p_i\in S$ and $g_j\in\R^d$ for the point $q_j\in T$.

Given that $S$ represents a sampling of the source domain, we wish to find the flow in space of each point in $S$. We denote by $f_i \in \R^3$ the flow vector of point $p_i$ which is shifted towards $p_i+f_i$ in the target domain. Note that we do not learn a correspondence between the source points and the target points but a flow representation of each point on the source.
 Due to a potential occlusion, some points in the source may not appear in the target frame. We note the occlusion of a point $p_i$ in the source using a binary scalar $occ_i\in\{0, 1\}$, where 0 means occluded and 1 means non-occluded. Our goal is to find the scene flow $\{f_i\}_{i=1}^{n_1}$ and occlusion label $\{occ_i\}_{i=1}^{n_1}$ for every point in the source.


\section{Architecture}
Inspired by the architecture of~\cite{wu2020pointpwc}, our network utilizes a feature pyramid structure and uses the point cloud from two different time frames as its inputs, where each point can have a rich feature vector such as color or normal to the surface. In the examples shown in this paper, we use the RGB color as our input point feature. See Figure \ref{fig:full_arch} for network architecture. In each pyramid level, we first apply a backward warping of the target point cloud $T$ towards the source $S$ by using the upsampled flow from the previous level. Then, by using the features from the point clouds and the upsampled occlusion mask from the previous level, we construct our cost volume for each point in $S$. The cost volume is a widely used concept in stereo matching~\cite{Sun2018:Model:Training:Flow,5995372}. It stores the point-wise matching cost and measures the correlation between the different frames. Finally, we predict the finer flow and mask by using the cost volume, features from $S$, upsampled flow and mask.

\noindent
\textbf{Feature Pyramid Structure}. In order to extract the semantically strong features for the accurate flow and occlusion mask prediction, we construct a 4-level pyramid of features with the input at the top (\textit{zeroth}) level. For each pyramid level \textit{l}, we downsample the point clouds for the coarser level (\textit{l+1}) by using the farthest point sampling (FPS)~\cite{qi2017pointnetplusplus}. Followed by a PointConv~\cite{wu_qi_fuxin_2019} operation, we create and increase the number of the features for each downsampled point. The finer prediction of flow and mask at each level are made by using the upsampled prediction from its coarser level (except for the bottom level).

\noindent
\textbf{Warping}. At each pyramid level, we first do a backward warping of the target points towards the source by using the upsampled scene flow from the previous coarser level. We use the same Upsample layer as in~\cite{wu2020pointpwc}. Since the warping layer brings the target “closer” to the source, neighborhood searching of the target around the source point would be more accurate during the cost volume construction. Denote the upsampled flow from the coarser layer as $\{ f_i^{up}\}_{i=1}^{n_1}$. Inside the warping layer, we first do a forward warping from the source to the target:

\begin{equation}
S_w = \{p_{w,i}=p_i + f_i^{up} \}_{i=1}^{n_1}
\label{foward warp}
\end{equation}
For each point $q_j$ in the target $T$, we compute its backward flow by using the weighted average of the upsampled flow:

\begin{equation}
f_j^b =\frac{\sum_{p_i\in N_{Sw}(q_j)}w(p_i, q_j)\times(-f_i^{up})}{\sum_{p_i\in N_{Sw}(q_j)}w(p_i, q_j)}
\label{backward flow}
\end{equation}
\noindent
Where $N_{Sw}(q_j)$ is the K nearest neighbor (k-NN) of $q_j$ on $S_w$, weight $w(p_i, q_j)=\frac{1}{d(p_i, q_j)}$ is simply the inverse of the euclidean distance between $p_i$ and $q_j$. Finally, the warped target point will be the element-wise addition of the backward flow and itself:

\begin{equation}
T_w = \{q_{w,j}=q_j + f_j^b \}_{j=1}^{n_2}
\label{backward warp}
\end{equation}

\begin{figure}
\begin{center}
\vspace{-16pt} 
\includegraphics[width=0.5\textwidth]{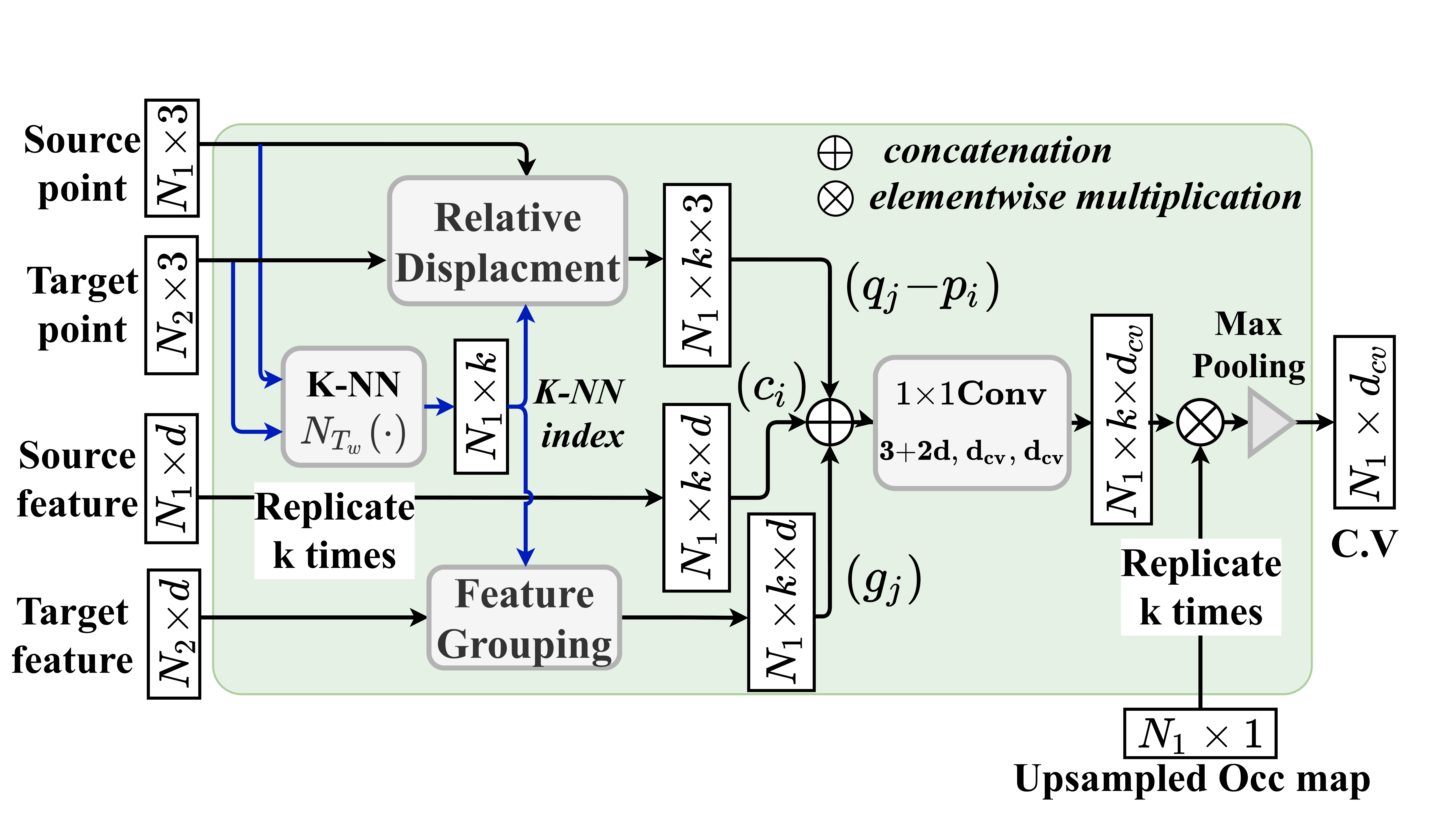}
\end{center}
   \vspace{-21pt} 
   \caption{\textbf{Cost Volume Layer}. For each point in the source, we first find its k-NN point in the target. Then, we group the relative displacement ($q_j-p_i$) and the features ($g_j$) of the neighborhood. After we calculate the matching cost, we apply the occlusion masking and the Max-pooling to construct the Cost Volume.}
\label{fig:cv}
\end{figure}

\noindent
\textbf{Cost volume with Occlusion mechanism}. Traditionally, occlusions play an essential role in the scene flow estimation on the 2D stereo frames. When it comes to the 3D point clouds, occlusion issues still exist due to the motion of the object and the camera position. The main impact of the occlusion is on the cost volume since the matching cost for the occluded point is not available. Similar to the images, the occlusion in the source point cloud relative to the target can be modeled as a map: $OCC_{S{-}T}:S\rightarrow[0,1]$ where 0 stands for the occluded point, 1 stands for the non-occluded. FlowNet3D~\cite{liu:2019:flownet3d} uses a flow embedding layer to aggregate the features and spatial relationships for each neighbor in the target around the source. Since their model only finds the neighboring points within a certain radius, it is somehow robust to the occlusion as the relative displacement between the occluded point and target is usually large. PointPWC-Net~\cite{wu2020pointpwc} suggests a novel cost volume that can aggregate the features of both input point cloud in a patch-to-patch manner. However, for the occluded regions in the source, this feature aggregation operation can be incorrect since they do not have a correspondence in the target frame. Inspired by PWOC-3D~\cite{saxena2019pwoc}, we suggest a novel occlusion mechanism that helps the construction of our cost volume. 

One of the critical components of cost volume is the matching cost. It measures the similarity between the source point and the target point. Since we believe that the correlation between the points is highly related to their features and relative displacement, for the \textit{non-occluded} point $p_i$, the matching cost between $p_i$ and $q_j$ is calculated by
\begin{equation}
cost(p_i,q_j )=h(c_i,g_j,q_j-p_i)
\label{non-occ cost}
\end{equation}
\noindent
Where $h(\cdot)$ is simply a concatenation of its input followed by $1{\times}1$ convolution layers, $c_i$ and $g_j$ are the corresponding features of $p_i\in S$ and $q_j\in T_w$. When it comes to the \textit{occluded} points $p_i$ , we expect to get a matching cost of 0, as they do not have a correspondence in the target frame. As shown in Figure~\ref{fig:cv}, by using our definition of occlusion map, we can calculate our matching cost of $p_i$ with $q_j$ as:

\begin{equation}
cost(p_i,q_j)=OCC_{S-T}(p_i)h(c_i,g_j,q_j-p_i)
\label{cost}
\end{equation}
In our case, we use the upsampled predicted occlusion mask from its coarser layer as the occlusion map in Eq.~\ref{cost}.

After we calculate the matching cost, we can aggregate them to form the cost volume. Theoretically, we can use all possible pairs of $(p_i,q_j)$ in our calculation, but this is inefficient in terms of the computation. With the help of the Warping layer, we can assume that the correct corresponding point pairs between source and target are relatively close to each other. For this reason, we only aggregate the matching cost of the nearest target neighbor for every point in the source. It can be summarized in the following form:
\begin{equation}
CV(p_i)=\underset{q_j\in N_{Tw}(p_i)}{Aggregation}\{cost(p_i,q_j )\}
\label{cost_volume}
\end{equation}
where $N_{Tw}(p_i)$ is the nearest neighborhood of the source point $p_i$ in the warped target $T_w$.

In the cost volume layer of~\cite{wu2020pointpwc}, they use a learnable weighted sum based on the relative distance as the aggregation function to calculate their Point-to-Patch cost. 
This implies that the proportion of the matching cost between $(p_i,q_j)$ in $CV(p_i)$ only depends on their relative displacement $(q_j-p_i)$. However, in many cases, the correlation between the points depends on their features but not their relative displacement, the correct corresponding pairs can have less contribution to the cost volume by using this aggregation design. In our work, we decide to use the max-pooling to aggregate the matching cost. The intuition is that, to make an accurate prediction of the flow and mask, the model needs the matching cost of the correct corresponding pairs to have the highest contribution in the cost volume. Using max-pooling can force their matching cost to have the highest value among the neighborhood $N_{Tw} (p_i)$ during training. This choice of design also agrees with our definition of the matching cost above. To summarize, we calculate the cost volume for every point $p_i$ by using the following equation:

\begin{equation}
CV(p_i)=\underset{q_j\in N_{Tw}(p_i)}{MAX}\{cost(p_i,q_j)\}
\label{cost volume}
\end{equation}

\begin{figure}
\begin{center}
\vspace{-16pt}
\includegraphics[width=0.5\textwidth]{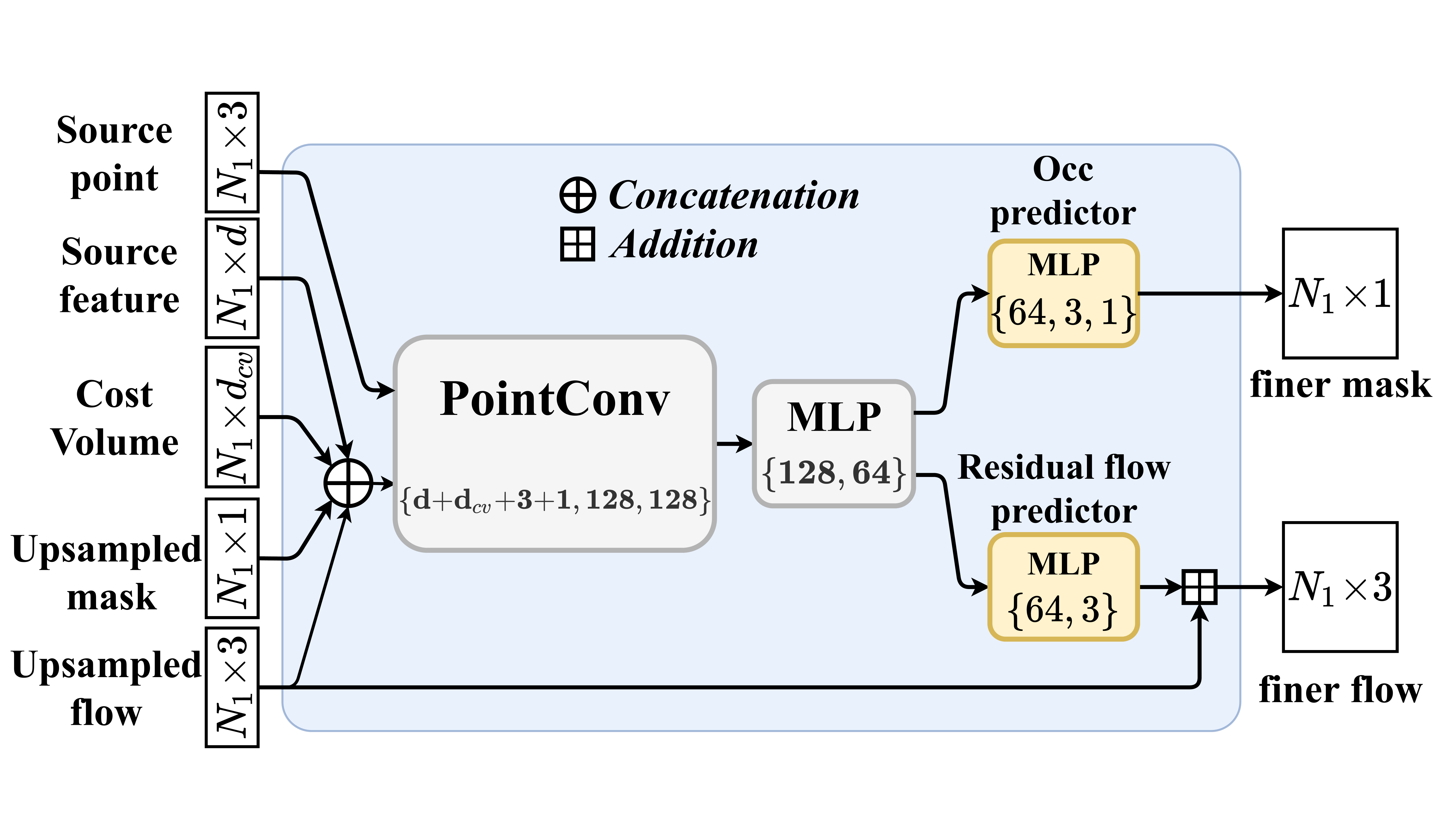}
\end{center}
   \vspace{-31pt}
   \caption{\textbf{Predictor Layer}. Our Predictor layer takes several inputs and produces the scene flow and occlusion mask at current level. These outputs will be upsampled and used as one of the inputs in the the next pyramid level.}
\label{fig:predictor}
\end{figure}

\noindent
\textbf{Predictor Layer}. In order to make the final prediction of the flow and occlusion mask at each pyramid level, we use a predictor layer. As shown in Figure~\ref{fig:predictor}, this layer contains a feature propagation module followed by two predictor branches. In the feature propagation module, we first concatenate all its inputs along the feature dimension. Then by using several PointConv and Multilayer perceptron (MLP), we generate the final features for the flow and mask prediction. The inputs of the feature propagation module are the features of the source, masked cost volume described above, upsampled flow and upsampled occlusion mask. After the feature propagation layer, we connect a flow predictor and occlusion predictor in parallel. Since we believe that the scene flow and occlusion are highly related to each other, we decide to use the shared input features for the two branches. Our flow predictor consists of a single MLP layer such that the output tensor has a dimension of ($n_1$, 3). Unlike the PointPWC-Net~\cite{wu2020pointpwc}, our flow predictor only predict a \textit{residual} flow vector such that the final scene flow is the element-wise addition of the upsampled flow from the previous level and the residual flow for every point in the source. By using this residual flow design, we solve the scene flow estimation problem in an iterative approach and we get a stronger correlation between the consecutive pyramid levels. Shifting from multi-scale flow estimation to multi-scale residual improve the results significantly and we show that in the ablation study.

For the occlusion branch, we use a 2-layer MLP with leaky-ReLU activation in the middle to process the input features. We also connect a sigmoid activation layer in the end. This ensures the output to be a probability distribution with value in the range [0, 1].
\section{Loss functions}
We train our model in a supervised manner with the ground truth scene flow and occlusion mask. Since the existing scene flow dataset with real scans is too small for the training, we adopt a similar training scheme as in the previous work~\cite{liu:2019:flownet3d,wu2020pointpwc}. We first train our model with the synthetic data from the FlyingThings3D~\cite{mayer_ilg_hausser_fischer_cremers_dosovitskiy_brox_2016}, then we test it with the real LiDAR scans from the KITTI~\cite{menze_geiger_2015,menze_heipke_geiger_2015}. We show that OGSF-Net has the best generalization ability to the unseen data from KITTI in the experiment section. In order to predict both the scene flow and occlusion map, we use two loss terms to train our model.

\noindent
\textbf{Scene flow loss}. We use a similar loss function as in ~\cite{liu:2019:flownet3d} and ~\cite{wu2020pointpwc} for the flow estimation. Let $f_i^\prime$ be the ground truth flow and $f_i$ be the predicted flow for the point $p_i\in S$. Let $occ_i^\prime$ be the ground truth occlusion label for $p_i$ with the value in $\{0, 1\}$. We use a multi-level loss for the flow as below

\begin{equation}
F_{loss}(\Theta)= \sum\limits_{l=0}^{3}\alpha_l\sum\limits_{p_i\in S_l}occ_i^\prime\big\|f_i-f_i^\prime \big\|_2+\big\|f_i-f_i^\prime \big\|_2
\label{flow loss}
\end{equation}
Where $\Theta$ is the learnable parameters of OGSF-Net, $S_l$ is the sampled point cloud at pyramid level $l$, and $\alpha_l$ is the weight for each level. The $first$ term in the inner summation penalizes the $L_2$ norm of the errors in estimated flow for non-occluded regions. Since we also want to predict the flow for occluded regions, we add the $second$ term which penalizes the error for all points in every $S_l$ and it improves the performance through our experiments.

\noindent
\textbf{Occlusion loss}. At each pyramid level, we use the predicted occlusion map to construct our masked cost volume. It means accurate mask prediction is also important for flow estimation at each level. Let $occ_i^\prime$ be the ground truth occlusion label and $occ_i$ be the predicted label for the point $p_i\in S$. We use a similar occlusion loss as the flow loss:

\begin{equation}
O_{loss}(\Theta)= \sum\limits_{l=0}^{3}\beta_l\sum\limits_{p_i\in S_l}\|occ_i-occ_i^\prime\|
\label{occ loss}
\end{equation}

\noindent
The overall loss function we used is simply the combination of the flow and occlusion loss from each pyramid level:
\begin{equation}
L(\Theta)= F_{loss}(\Theta) + \lambda \cdot O_{loss}(\Theta)
\label{all loss}
\end{equation}
We use the $\lambda$ as a weight to control the balance between the flow loss and occlusion loss.


\begin{table*}[t!]
\centering
\begin{tabular}{@{}l|l|ccccc@{}}
\toprule
\multicolumn{1}{c|}{Dataset}    & \multicolumn{1}{c|}{Method} & EPE$_{full}{\downarrow}$        & EPE$\downarrow$             & ACC$_{05}{\uparrow}$	         & ACC$_{10}{\uparrow}$	        & Outliers$\downarrow$        \\ \midrule
\multirow{6}{*}{Flyingthings3D} & ICP~\cite{10.1117/12.57955}   & 0.5048          & 0.4848          & 0.1215          & 0.2558          & 0.9441          \\
                                & FlowNet3D~\cite{liu:2019:flownet3d}  & 0.2119          & 0.1577          & 0.2286          & 0.5821          & 0.8040          \\
                                & HPLFlowNet~\cite{gu_wang_wu_lee_wang_2019} & 0.2012          & 0.1689          & 0.2629          & 0.5745          & 0.8123          \\
                                & FLOT$(K=1)$~\cite{puy20flot}  & 0.2502          & 0.1530          & 0.3965          & 0.6608          & 0.6625          \\
                                & PointPWC-Net~\cite{wu2020pointpwc} & 0.1953          & 0.1552          & 0.4160          & 0.6990          & 0.6389          \\ 
                                \cmidrule(l){2-7} 
                                & Ours                        & \textbf{0.1634} & \textbf{0.1217} & \textbf{0.5518} & \textbf{0.7767} & \textbf{0.5180} \\ \midrule
\multirow{7}{*}{KITTI}          & ICP~\cite{10.1117/12.57955}    & 0.3801          & -               & 0.1038           & 0.2913          & 0.8307          \\
                                & FlowNet3D~\cite{liu:2019:flownet3d}  & 0.1834          & -               & 0.0980          & 0.3945          & 0.7993          \\
                                & HPLFlowNet~\cite{gu_wang_wu_lee_wang_2019} & 0.3430          & -               & 0.1035          & 0.3867          & 0.8142          \\
                                & FLOT$(K=1)$~\cite{puy20flot}   & 0.1303           & -               & 0.2788          & 0.6672          & 0.5299          \\
                                & PointPWC-Net~\cite{wu2020pointpwc} & 0.1180          & -               & 0.4031          & 0.7573          & 0.4966          \\ \cmidrule(l){2-7} 
                                & Ours($without$ $ft$)                &0.0751  & -    & 0.7060  & 0.8693  & 0.3277 \\ 
                                & Ours($with$ $ft$)                      &\textbf{0.0333}  & -    & \textbf{0.8913}  &\textbf{0.9517}  &\textbf{0.1915} \\ \bottomrule
\end{tabular}
\vspace{-9pt}
\caption{\textbf{Performance on Flyingthings3D and KITTI}. All the models in the table are trained on the occluded Flyingthings3D using 8192 points. We test it on KITTI (with occlusion) using 8192 points from each frame $without$ any fine-tuning. Notice that we outperforms all other methods by a large margin. In the last column, we also present our fine-tuned results on KITTI.}
\label{tab:epe}
\end{table*}

\section{Experiments}
In this section, firstly, we compared the performance of our OGSF-Net with previous work on the FlyingThings3D~\cite{mayer_ilg_hausser_fischer_cremers_dosovitskiy_brox_2016} synthetic dataset on several evaluation metrics. Without any fine-tuning, we also test our model's generalization ability on the real scans from KITTI~\cite{menze_geiger_2015,menze_heipke_geiger_2015}. By further fine-tuning on KITTI, we show improvements in the results and present visualization on KITTI. In the previous works, there are two versions of FlyingThings3D and KITTI that have been proposed. The first one is suggested by~\cite{gu_wang_wu_lee_wang_2019}, where the occluded point is removed from the processed point cloud and many difficult examples in the Flyingthings3D have been removed. The second version is suggested by FlowNet3D~\cite{liu:2019:flownet3d}. The occluded region remains and the occlusion map for FlyingThings3D is provided. Since our work is highly related to the occlusion, we adopt the FlyingThings3D and KITTI proposed by~\cite{liu:2019:flownet3d}, which is more challenging than the first version. Secondly, in the ablation study, we test our design choices and show the effectiveness of all the novel components in our work. Finally, we evaluate our occlusion estimation. To the best of our knowledge, we are the first one to evaluate the occlusion on scene flow estimation on point clouds. We present here state-of-the-art results compared to those of previous reported methods.

\noindent
\textbf{Evaluation Metric}. We first adopt the four evaluation metrics used in~\cite{gu_wang_wu_lee_wang_2019,liu:2019:flownet3d, wu2020pointpwc,puy20flot}: averaged end point error (EPE); two accuracy measurement with a different threshold on EPE; outlier ratio with a threshold on the EPE. In ~\cite{liu:2019:flownet3d, puy20flot}, the above metrics are evaluated on the non-occluded points only, while in our work, we evaluate results for all the points, include occluded and non-occluded ones. The details of the evaluation metrics are as follow:

\renewcommand{\labelitemi}{$\star$}
\begin{itemize}
\item $EPE_{full}$(m): $\big\|f_i-f^{\prime}_{i}\big\|_2$ averaged over \textbf{all} $p_i\in S$.
\item $EPE$(m): $\big\|f_i-f^{\prime}_{i}\big\|_2$ averaged over all \textbf{non occluded} points.
\item $ACC_{05}$: percentage of points whose $EPE_i$ $<$ 0.05m or $EPE_i$ $/$ $\big\|f^{\prime}_{i}\big\|_2$ $<$ 5\%
\item $ACC_{10}$: percentage of points whose $EPE_i$ $<$ 0.1m or $EPE_i$ $/$ $\big\|f^{\prime}_{i}\big\|_2$ $<$ 10\%
\item $Outlier$: percentage of points whose $EPE_i$ $>$ 0.3m or $EPE_i$ $/$ $\big\|f^{\prime}_{i}\big\|_2$ $>$ 10\%

\end{itemize}
\textbf{Implementation Details}. Our OGSF-Net utilizes the same feature pyramid structure as in~\cite{wu2020pointpwc} to process the input point clouds, while the number of points we used in each downsampled point cloud is $[2048,512,256,128]$. We choose the weight $\alpha$ in the Eq.~\ref{flow loss} to be $\alpha$ = $[\alpha_l]_{l=0}^{3}$ = $[0.02,0.04,0.08,0.16]$. The weight $\beta$ in Eq.~\ref{occ loss} is set to be $\beta_l$ = $1.4\alpha_l$ for every pyramid level $l$. The number of features $d$ at each level is set to be $[64,96,192,320]$ and $d_{cv}$ at each level is $[32, 64, 128, 256]$. All hyper-parameters are selected according to the validation set of Flyingthings3D. We trained our model with 2$\times$GTX2080Ti GPU on FlyingThings3D with batch size of 8 and 120 training epochs, and it took one day to train. We start with a learning rate of $0.001$ and reduce it after every 10 epochs with a decay rate $0.85$. We further reduce the decay rate to $0.8$ after 75 epochs. The balancing weight $\lambda$ is 0.3 initially. In order to improve the occlusion accuracy, we increase the $\lambda$ gradually to $0.6$ in the first 45 epochs.

\begin{figure*}[t]
\begin{center}
\vspace{-2pt}
\includegraphics[width=1.0\textwidth]{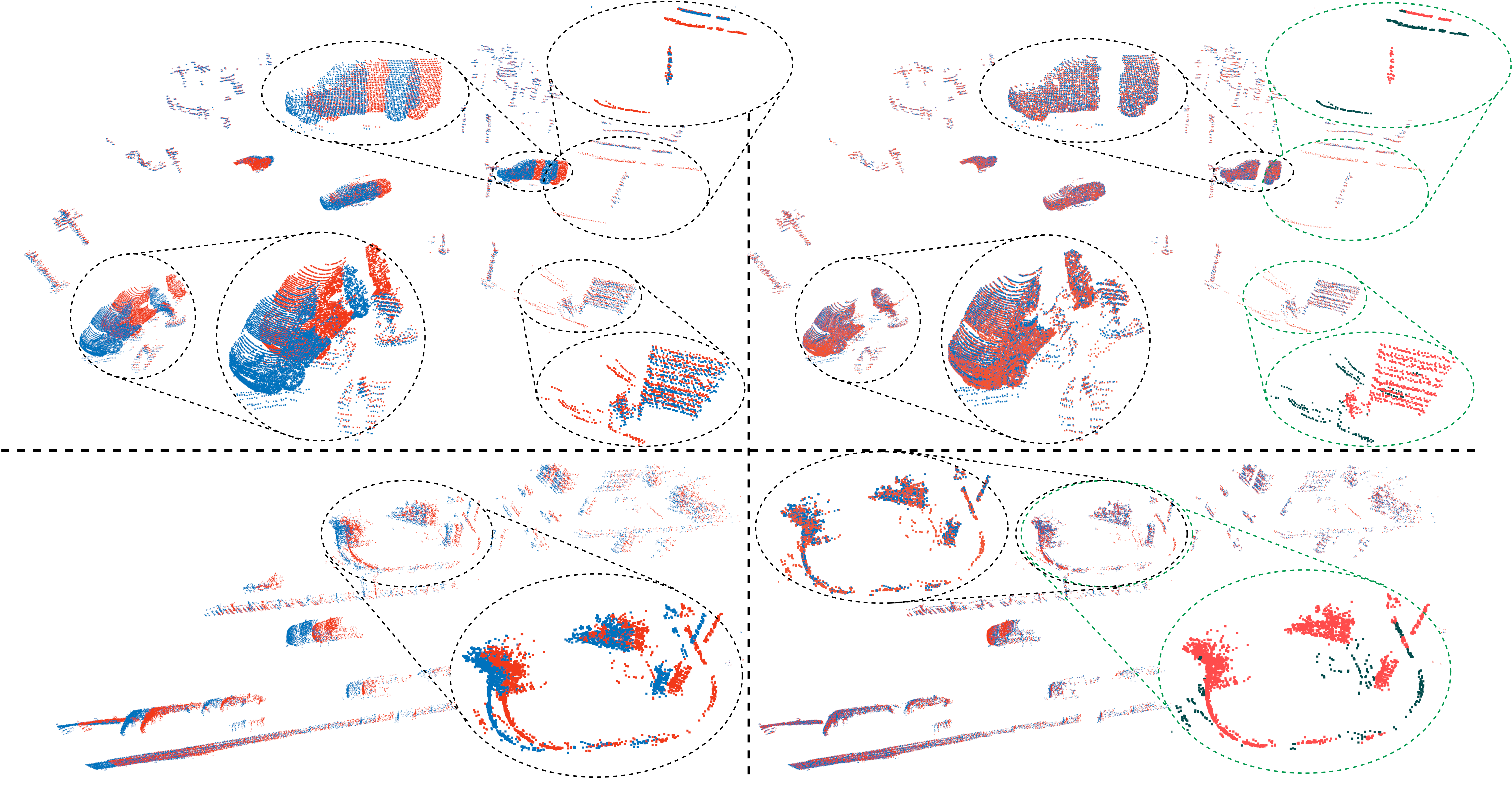}
\end{center}
   \vspace{-20pt}
   \caption{\textbf{Visualization on KITTI Scene Flow 2015}. For the images on the left, we show the source (red) and target (blue) point cloud on the same 3D space before the alignment. For the images on the right, we align the source towards the target by using the predicted flow from OGSF-Net (source+scene flow). The zoom-in view for the region circled by black is shown. We also provide the zoom-in detail of our predicted occlusion map for the region circled by green. We can see that our OGSF-Net can predict the map for the occluded points (black) and non-occluded points(red) correctly, it can also estimate the accurate flow for both occluded and non-occluded regions.}
\label{fig:kitti}
\end{figure*}

\subsection{Evaluation on Flyingthings3D}
Since the acquisition of dense flow and occlusion mask from the real scene is difficult, to the best of our knowledge,  there is no real-world large-scale scene dataset published with ground truth flow and mask. Thus, by following the similar evaluation process in ~\cite{liu:2019:flownet3d, puy20flot, wu2020pointpwc, gu_wang_wu_lee_wang_2019}, we trained our model on the synthetic FlyingThings3D~\cite{mayer_ilg_hausser_fischer_cremers_dosovitskiy_brox_2016} dataset. As mentioned before, we use the same dataset suggested by ~\cite{liu:2019:flownet3d}, it contains 20000 pairs of the point cloud in the training set and 2000 in the test set. 

Since both the ground truth scene flow and occlusion mask are provided in this dataset, we use the loss function in 
Eq.\ref{all loss} to train our model.

The detailed comparison results are shown in Table \ref{tab:epe}. We compared our model with the previous state-of-the-art methods on point cloud scene flow estimation. All the methods were trained on Flyingthings3D proposed by~\cite{liu:2019:flownet3d}, we use $n_1=n_2=8192$ points for each point cloud for the training and evaluation. It is clear to see that our method outperforms the previous work in all evaluation metrics. As mentioned in the related work, when compared the numbers in Table~\ref{tab:epe} to the reported results in their own paper, we can see that the performance of~\cite{wu2020pointpwc,gu_wang_wu_lee_wang_2019} is highly degraded due to the existence of occlusion in the input. Notice that the performance of FlowNet3D~\cite{liu:2019:flownet3d} and  FLOT~\cite{puy20flot} is acceptable on the EPE, but they perform much worse on the EPE$_{full}$. This is because they removed the errors for the occluded region in their loss function and they are not able to predict the flow for the occluded points.


\begin{figure*}
\begin{center}
\includegraphics[width=1.0\textwidth]{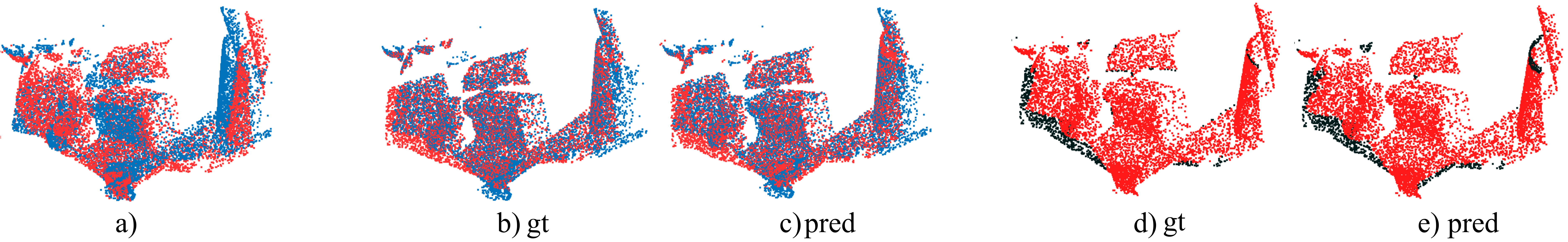}
\end{center}
    \vspace{-19pt}
   \caption{\textbf{Flow/Occlusion Visualization on Flyingthings3D}. An example from the test set of Flyingthings3D. a) shows the source (red) and target (blue) frames, b) and c) show the alignment results by using the ground truth and predicted flow, d) and e) show the ground truth and predicted occlusion map where the non-occluded region is marked in red and the occluded region is marked in black.} 
\label{fig:occ}
\end{figure*}

\subsection{Evaluation on KITTI} In order to test the generalization ability on real scans, we first trained our model on Flyingthings3D then test it on all the 150 examples with $n_1=n_2=8192$ points from KITTI Scene Flow 2015~\cite{menze_geiger_2015,menze_heipke_geiger_2015} without any fine-tuning. Since they do not provide the ground truth occlusion map for the source, we cannot evaluate  the EPE on the KITTI.

As shown in Table~\ref{tab:epe}, our model has the best generalization ability compared to the previous work. On the last row in the Table~\ref{tab:epe}, we split the data to 100 training samples for fine-tuning and 50 samples for test, we show a further improvement in the performance. Since there is no ground truth occlusion mask, we only use $\sum\limits\alpha_l\sum\limits\big\|f_i-f_i^\prime \big\|_2$ ($second$ term of the $F_{loss}(\Theta)$) as the loss function for the fine-tuning.

\subsection{Ablation Study} We performed several ablation studies to validate our model's design choices, occlusion guided mechanism, and loss functions. In Table~\ref{tab:abl} (a), we report the EPE of a different combination of the design choices on the Flyingthings3D dataset. When we use Max-pooling to aggregate the matching cost in the Cost Volume layer, we obtain significantly better results in terms of the EPE. By further using our residual flow prediction design instead of the full scene flow prediction, we got a $19\%$ improvement in the performance. In the last two rows, we show that our model's performance on the occluded dataset improved by a large margin by utilizing the occlusion estimation mechanism. In Table~\ref{tab:abl} (b), we train our model using the different loss functions and present the EPE and EPE$_{full}$ on the Flyingthings3D and KITTI respectively. 
As shown in the bottom row, OGSF-Net can distinguish between the occluded and non-occluded regions by training with the occlusion loss. It improves the performance on the Flyingthings3D and we got a better generalization ability on KITTI.

\begin{table}
\begin{subtable}{0.5\textwidth}

\begin{tabular}{@{}c|cc|c|c@{}}
\toprule
Aggregation  & \begin{tabular}[c]{@{}c@{}}Occ.\\ Predictor\end{tabular} & \begin{tabular}[c]{@{}c@{}}Masked\\ C.V\end{tabular} & \begin{tabular}[c]{@{}c@{}}Flow\\ branch\end{tabular}     & EPE$\downarrow$    \\ \midrule
Weighted sum & \XSolidBrush                                             & \XSolidBrush                                          & Full     & 0.1610 \\
Weighted sum & \Checkmark                                               & \XSolidBrush                                          & Full     & 0.1541 \\
Weighted sum & \Checkmark                                               & \Checkmark                                            & Full     & 0.1512 \\ \midrule
Max-Pooling  & \XSolidBrush                                             & \XSolidBrush                                          & Full     & 0.1503 \\
Max-Pooling  & \XSolidBrush                                             & \XSolidBrush                                          & Residual & 0.1304 \\ \midrule
Max-Pooling  & \Checkmark                                               & \Checkmark                                            & Residual & \textbf{0.1217} \\ \bottomrule
\end{tabular}

\caption{Design choice}
\end{subtable}%
\vspace{0.2cm}

\begin{subtable}{0.5\textwidth}
\begin{center}

\begin{tabular}{@{}cc|c|c@{}}
\toprule
FLow loss & \multicolumn{1}{c|}{Occlusion loss} & \multicolumn{1}{c|}{Flyingthings3D} & KITTI  \\ \midrule
\Checkmark         & \multicolumn{1}{c|}{\XSolidBrush}              & \multicolumn{1}{c|}{0.1337}         & 0.0794 \\
\Checkmark         & \multicolumn{1}{c|}{\Checkmark}              & \multicolumn{1}{c|}{\textbf{0.1217}}         & \textbf{0.0751}  \\ \midrule

\end{tabular}
\end{center}
\vspace{-15pt}
\caption{Loss function}
\end{subtable}%

\vspace{-7pt}
\caption{\textbf{Ablation Studies for the model design}.$(a)$ we show the different combination of design choices, and ours can get the best performance.$(b)$ by training with the occlusion, we can get a much better generalization on the real scans from KITTI.}
\label{tab:abl}
\end{table}

\subsection{Occlusion Estimation} 
An accurate occlusion prediction is important for our occlusion-guided mechanism and important for some applications like 3D object reconstruction.
In this section, we evaluate the performance of occlusion estimation on the Flyingthings3D only as there is no other public dataset on point cloud that provides the ground truth occlusion mask. We use the standard occlusion estimation metrics, accuracy and F1-score, as our evaluation metrics. We first convert the predicted occlusion probabilities to the label $\{0,1\}$ using threshold value 0.5. Then, we measure the two metrics and we get 94.91\% and 0.824 respectively. We also showed some visualization of the occlusion estimation results in Fig~\ref{fig:kitti} and ~\ref{fig:occ}.




\begin{table}[t]
\begin{center}
\begin{tabular}{|c|c|c|c|c|c|}
\hline
\multirow{2}{*}{Method} & \multicolumn{5}{c|}{Threshold for Outlier (m)}                                 \\ \cline{2-6} 
                        & 0.1            & 0.2           & 0.3           & 0.4           & 0.5           \\ \hline
FlowNet3D~\cite{liu:2019:flownet3d}               & 67.87          & 29.15         & 14.41         & 7.83         & 4.46          \\ \hline
FLOT~\cite{puy20flot}                    & 37.34          & 16.69         & 9.22          & 5.37          & 3.37 \\ \hline
Ours                    & \textbf{13.82} & \textbf{6.54} & \textbf{4.78} & \textbf{3.85} & \textbf{3.27}          \\ \hline
\end{tabular}
\end{center}
\vspace{-17pt}
\caption{\textbf{Outlier ratio}. We measure the outlier ratios with different threshold values. We only compared our model with the FlowNet3D and FLOT as they are the only models trained and tested with occluded data in their works.}
\label{tab:out}
\end{table}

\subsection{Outlier ratios}In the scene flow estimation, outlier ratios are important as they measure the robustness of the model. In Table~\ref{tab:out}, we show outlier ratios on KITTI Scene Flow 2015~\cite{menze_geiger_2015,menze_heipke_geiger_2015} with different threshold values for different models. We calculate the ratio by simply finding the percentage of the point whose EPE$_{full}$ is greater than the given threshold. As we can see, the performance of~\cite{puy20flot} and ours is much better than~\cite{liu:2019:flownet3d}. For all the threshold values from 0.1 to 0.5, our model has the smallest outlier ratio compared to the FlowNet3D~\cite{liu:2019:flownet3d} and FLOT~\cite{puy20flot}.

\section{Conclusion}In this paper, we suggest a deep neural network called OGSF-Net that can jointly estimate the scene flow and occlusion map directly from the point cloud data. We are the first to introduce the idea of occlusion estimation on the point cloud scene flow estimation, and by using our masking operation inside the Cost Volume layer, we show a significant improvement in the flow accuracy. Our occlusion guided flow estimation not only provides an additional layer of information but outperforms previously reported state-of-the-art models by a large margin, on multiple datasets and for different metrics.

{\small
\bibliographystyle{ieee_fullname}
\bibliography{version2}
}

\end{document}